\documentclass[10pt,twocolumn,letterpaper]{article}

\usepackage[pagenumbers]{cvpr} %

\usepackage{amsthm}
\usepackage{multirow}
\usepackage{colortbl}
\usepackage{algorithm}
\usepackage{algpseudocode}

\newcommand{\method}{Mem2Fair\xspace}
\newcommand{\pmstd}[1]{{\scriptsize\,$\pm$\,#1}}
\newcommand{\na}{\textnormal{--}}

\theoremstyle{plain}
\newtheorem{theorem}{Theorem}
\theoremstyle{definition}
\newtheorem{assumption}{Assumption}

\definecolor{cvprblue}{rgb}{0.21,0.49,0.74}
\usepackage[pagebackref,breaklinks,colorlinks,allcolors=cvprblue]{hyperref}

\def\confName{CVPR}
\def\confYear{2027}

\title{\method: Online Fairness Control for Text-to-Video Generation via Fair Memory}
\author{%
Pengyu Chen\textsuperscript{1}\thanks{Equal contribution.}, Jiaming Zhang\textsuperscript{1}\footnotemark[1], Xiaohua Feng\textsuperscript{1}, Yuyuan Li\textsuperscript{2}, Chaochao Chen\textsuperscript{1}\\
\textsuperscript{1}Zhejiang University \quad \textsuperscript{2}Hangzhou Dianzi University
}

\begin{document}
\maketitle
\begin{abstract}
\label{abstract}
Text-to-video (T2V) generation models can produce systematically unfair demographic outcomes when prompts leave protected attributes unspecified. For example, prompts describing nurses without specifying gender may disproportionately generate videos depicting female nurses.
Mitigating unfairness in T2V generation is particularly challenging: training-based methods that modify model internals require architecture-specific access and risk disrupting temporal consistency, while sampling-based methods often incur costly pre-deployment video generation.
We present the first conditional online approachability formulation of attribute-level fairness for T2V generation. Building on this formulation, we introduce \method, a plug-and-play framework that maintains concept-conditioned Fair Memory. Using generated videos at inference time as feedback, \method selects prompt-level attribute interventions based on the deviation from each concept's prescribed fairness target, without training or separate pre-deployment sampling.
Experiments demonstrate that \method consistently improves fairness over the unmodified generator and existing baselines across multiple demographic attributes, prompt settings, and target distributions, while maintaining comparable prompt–video alignment and video quality. 
We further demonstrate its effectiveness on commercial black-box T2V models.
\end{abstract}

\section{Introduction}
\label{sec:intro}

Text-to-video (T2V) models have developed rapidly in recent years, as illustrated by recent large-scale systems such as Wan 2.2, Seedance 2.0 and UniVideo~\citep{wan2025,seedance2026seedance20advancingvideo,wei2026univideo}. 
As these models become increasingly capable and widely deployed, ensuring fairness in their generated content has become an important concern.
Biases embedded in large-scale training data and complex model architectures can be reflected in generated content, leading attribute-unspecified prompts to produce systematically imbalanced outcomes across different attribute groups. Videos generated from prompts about specific concepts may exhibit imbalanced representation across gender, age, and race~\citep{luccioni2023stablebias,zhong2026fairt2v}.

These fairness concerns are well documented in text-to-image (T2I) models, where a growing body of work has proposed mitigation methods~\citep{fairmapping2025,fairgen2025iccv,fairimagen2025,lightfair2025}.
However, T2V fairness cannot be treated as a direct extension of T2I fairness because it presents three distinct challenges: 
\emph{(i) Temporal fidelity.}
Interventions that directly alter text embeddings, latent representations, or denoising trajectories may also affect temporal properties of generated videos, including inter-frame identity, motion, and consistency~\citep{geyer2024tokenflow}.
\emph{(ii) Dependence on model internals.}
Fine-tuning parameters, editing text encoders or attention modules, and changing denoising dynamics require access to architecture-specific components~\citep{shen2023finetuning,lightfair2025,teo2024fairqueuerethinkingpromptlearning,fairmapping2025,friedrich2023fairdiffusioninstructingtexttoimage}, limiting their applicability to advanced closed-source T2V systems.
\emph{(iii) Pre-deployment sampling cost.}
Many fairness methods rely on generated samples to train or construct their interventions. For T2V systems, producing these samples incurs substantial additional generation cost before deployment.

\begin{figure}[t]
    \centering
    \includegraphics[width=\columnwidth]{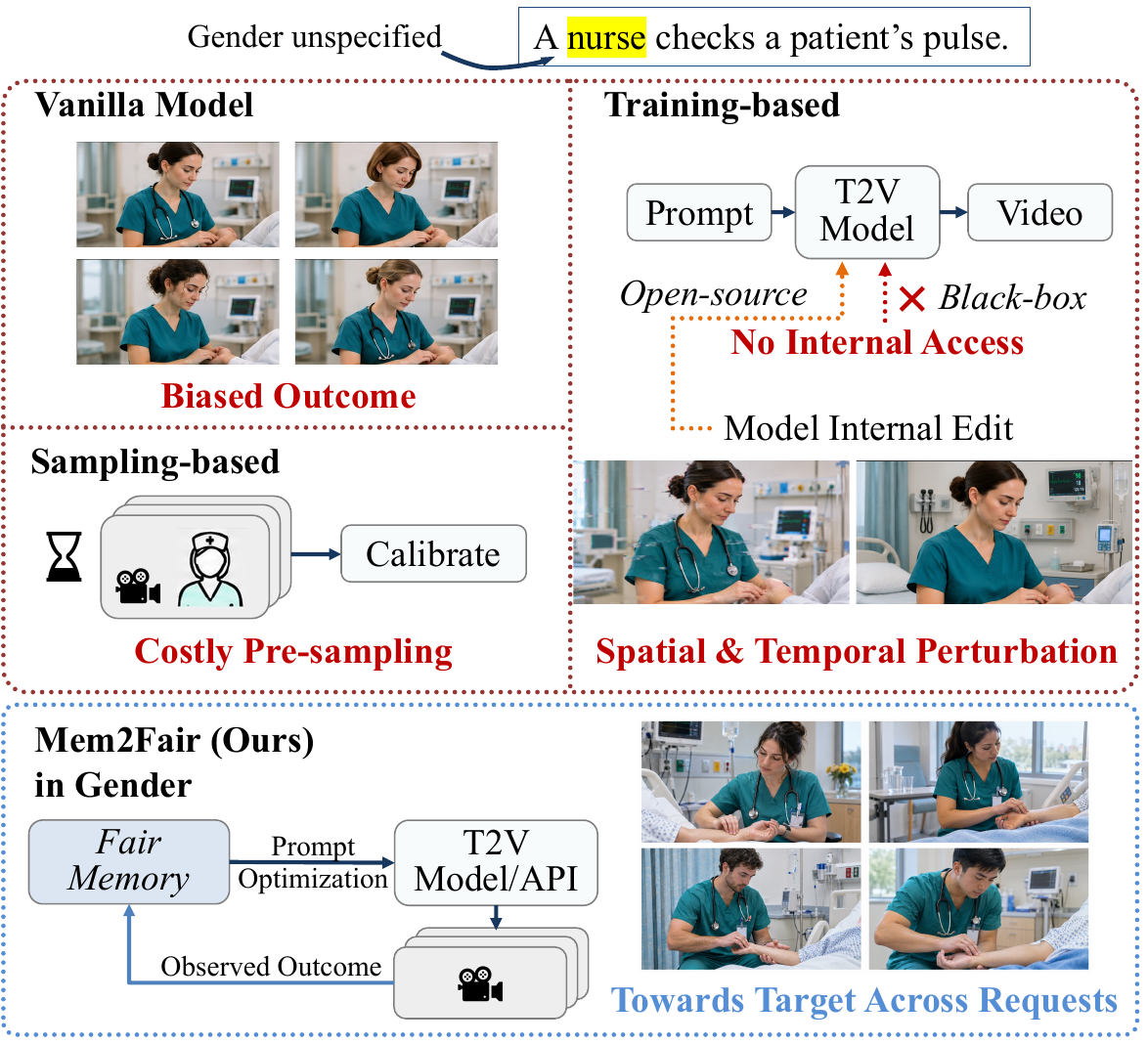}
    \caption{The nurse example illustrates demographic bias from a gender-unspecified prompt. Existing interventions face model-access constraints, fidelity disruption, or costly pre-sampling. \method uses Fair Memory and outcome feedback to guide prompt optimization, steering demographic proportions toward a prescribed target across requests.}\label{fig:motivation}
\end{figure}

To address these challenges, we formulate fair T2V generation as a conditional online approachability problem~\citep{blackwell1956,chzhen2021faironline}. This formulation treats fairness as an outcome-feedback problem in which generated videos provide observations that guide subsequent interventions for the corresponding semantic concept. By leveraging feedback collected during the sequential generation process, fairness control does not require exhaustive offline sampling across different prompts, events, and interventions, thereby avoiding the associated upfront sampling burden.

Building on this formulation, we propose \method, a plug-and-play closed-loop framework that interacts with T2V generators only through input prompts and generated videos. It maintains a separate Fair Memory for each semantic concept, tracking realized attribute outcomes. 
At each generation round, an online approachability-based controller uses the discrepancy between the concept's empirical outcome distribution and the prescribed target to select an attribute value, which is then incorporated into the input prompt through prompt optimization. After generation, an external attribute classifier observes the outcome and updates the corresponding memory, providing feedback for subsequent generations. This input/output-only design requires no access to T2V model internals, leaves the generator's temporal process unchanged, and supports both open- and closed-source systems.

Our main contributions are:
\begin{itemize}
    \item To the best of our knowledge, we are the first to formulate attribute-level fairness in T2V generation as a conditional online approachability problem, using realized video outcomes as feedback to guide sequential decisions.
    \item We develop \method, a plug-and-play framework that uses concept-conditioned Fair Memory to guide attribute selection and optimizes the input prompt to incorporate the selected attribute value, without modifying the video generator. We further establish a finite-sample bound on the expected concept-level fairness discrepancy.
    \item We evaluate \method across multiple T2V generators, protected attributes, prompt settings, and target distributions, demonstrating consistent fairness improvements with comparable video quality and prompt--video alignment. We further validate its applicability to commercial black-box T2V models.
\end{itemize}

\section{Related Work}
\label{sec:related-work}

\paragraph{Bias in Image and Video Generation and Mitigation Methods.}
Text-to-image (T2I) diffusion models systematically under-represent particular demographic groups and amplify social stereotypes, as quantified by auditing studies and fairness benchmarks~\citep{luccioni2023stablebias,bianchi2023faworldscale,cho2023dalleval,seshadri2024biasamplification,teo2023measuringfairness}. Existing mitigation methods intervene at different points in the generation pipeline. Model-adaptation methods fine-tune diffusion parameters or learn inclusive prompt tokens~\citep{shen2023finetuning,zhang2023itigen}; conditioning-space methods edit text embeddings or encoders~\citep{fairmapping2025,lightfair2025,fairimagen2025,gandikota2024uce}; and inference-time methods modify guidance, attention, latent directions, or sampling~\citep{friedrich2023fairdiffusioninstructingtexttoimage,teo2024fairqueuerethinkingpromptlearning,fairgen2025iccv,parihar2024balancing,choi2024fairsampling}. Some strategies can theoretically be adapted to T2V, but direct transfer may inherit challenges, including temporal-fidelity loss, dependence on model-specific internals, and high pre-deployment sampling costs when training or calibration is required.

Existing research on T2V fairness encompasses both fairness assessment and mitigation. Fairness assessment studies measure attribute-level biases in generated videos, including gender and racial associations~\citep{nadeem2025genderbiastexttovideogeneration,sun2026veatquantifiesimplicitassociations}, while VMDT~\citep{potter2025vmdtdecodingtrustworthinessvideo} evaluates fairness as part of broader model trustworthiness. At the training stage, controllable preference modeling has been explored by constructing gender-controlled preference data and training reward models to guide video-model alignment~\citep{cai2026preferencesprejudicerolealignment}. However, this strategy requires curated preference data and additional reward-model and video-model training. At inference time, explicit debiasing prompts can reduce measured attribute associations but may exacerbate them for certain occupations~\citep{sun2026veatquantifiesimplicitassociations}. FairT2V~\citep{zhong2026fairt2v} instead neutralizes prompt embeddings and applies interventions at selected denoising steps, requiring access to model internals.

\paragraph{Memory and Prompt Optimization for Generation.}
Memory has supported several objectives in generation, including preserving long-range video context and reusing information from earlier prompts. Context-as-Memory~\citep{yu2025contextmemorysceneconsistentinteractive} retrieves relevant past frames for scene-consistent long-video generation, while VideoMemory~\citep{zhou2026videomemoryconsistentvideogeneration} stores and updates entity representations to maintain consistency across narrative shots. For compositional T2V generation, TTOM~\citep{ttom2025} stores test-time optimization parameters under prompt-derived layout keys and reuses them to improve text-video alignment.

Prompt optimization is a practical way to improve conditional generation by rewriting or expanding user prompts to better match a generator's learned prompt distribution and to specify relevant semantic details. For instance, Promptist~\citep{hao2023optimizing} optimizes prompts toward those preferred by human users, aiming to improve text-to-image generation quality, while RAPO~\citep{gao2025rapo} adapts prompts to the text-to-video model's prompt distribution to improve the static and dynamic quality of generated videos.

\paragraph{Online Fairness and Blackwell Approachability.}
At deployment, T2V systems receive a stream of prompts. Fairness concerns the distribution of realized demographic outcomes across repeated generations, while each video provides only one observation. This connects the problem to online fairness, which studies cumulative outcome constraints~\citep{chzhen2021faironline}, and fairness over sequences of decisions~\citep{benade2018make,cembrano2026online}. Blackwell approachability provides a geometric framework for steering a running vector average toward a target set~\citep{blackwell1956}. These frameworks offer tools for analyzing distributional fairness under sequential feedback.

\section{Preliminaries}
\label{sec:preliminaries}

\paragraph{Fairness objective.}
Let the protected attribute take values in $\mathcal{A}=\{a_1,\dots,a_K\}$, and let $\Delta^{K-1}$ be the $(K-1)$-dimensional probability simplex. The prescribed target distribution is $\pi\in\Delta^{K-1}$. For each semantic concept $c$, let $p_c$ be the distribution of detected outcomes over attribute-unspecified generations. The ideal fairness objective is
\begin{equation}
\label{eq:fairness}
p_c = \pi, \qquad \forall c.
\end{equation}

\paragraph{Quality objective.}
Let $v_t$ and $v_t^{(0)}$ be the controlled and original videos for the same prompt $x_t$, and let $Q(v,x)$ measure video quality. Ideally, fairness interventions should preserve video quality within a small tolerance:
\begin{equation}
\label{eq:quality-goal}
\left|Q(v_t,x_t)-Q\bigl(v_t^{(0)},x_t\bigr)\right|\leq\epsilon_Q,
\qquad \forall t.
\end{equation}

\paragraph{Blackwell approachability.}
Fix a concept and let $t$ index its eligible generations. Write $z_t\in\Delta^{K-1}$ for the one-hot encoding of the observed outcome $Y_t$; its running average $\bar z_T$ is the empirical outcome distribution. Blackwell approachability seeks actions for which this average approaches a closed target set $\mathcal{S}\subseteq\Delta^{K-1}$~\citep{blackwell1956}:
\begin{equation}
\label{eq:average-payoff}
\bar z_T=\frac{1}{T}\sum_{t=1}^{T}z_t,
\end{equation}
\begin{equation}
\label{eq:approachability}
\operatorname{dist}(\bar z_T,\mathcal{S})\rightarrow0
\qquad\text{as }T\rightarrow\infty.
\end{equation}
For fairness, $\mathcal{S}$ is a target region around the prescribed distribution $\pi$.

\section{Method}
\label{sec:method}

\begin{figure}[t]
    \centering
    \includegraphics[width=\columnwidth]{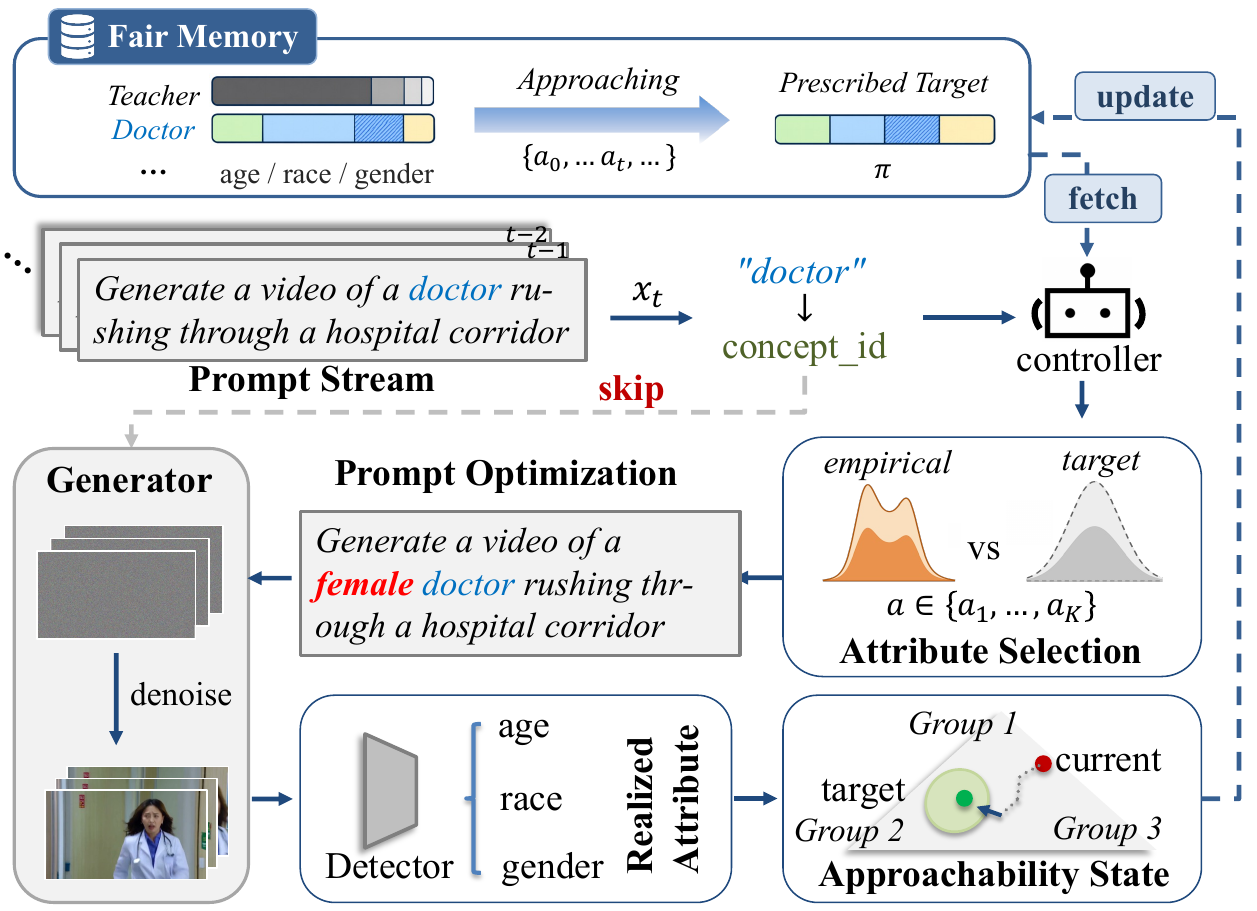}
    \caption{Overview of \method. Concept-conditioned Fair Memory guides fairness-aware prompt optimization through attribute selection and prompt optimization. Detected outcomes update the corresponding memory for subsequent requests.}
    \label{fig:method-overview}
\end{figure}

As illustrated in Figure~\ref{fig:method-overview}, \method performs fairness-aware prompt optimization using concept-conditioned Fair Memory. The memory guides attribute selection, and an LLM prompt optimizer incorporates the selected value into the input prompt while preserving its concept and context. After generation, the detected attribute updates the corresponding memory, closing the loop for subsequent requests.

\subsection{Concept-Conditioned Fair Memory}
\label{sec:fair-memory}
At request $t$, the prompt $x_t$ maps to a concept $c_t=C(x_t)$. We maintain a separate state for each concept because pooling outcomes can let frequent concepts mask bias in rarer ones. We only intervene on prompts that leave the protected attribute unspecified, while prompts that explicitly specify an attribute value are left unchanged.

An outcome is valid if the detector returns an attribute label; cases with no detected face or failed classification are excluded from Fair Memory updates. Let $\mathcal I_t^c$ index the attribute-unspecified requests for concept $c$ with valid detected outcomes up to time $t$, and let $N_t^c=|\mathcal I_t^c|$. For detected outcome $Y_s\in\mathcal A$, write $h(Y_s)$ for its one-hot encoding, with $h(a_k)=\mathbf e_k$. The empirical outcome distribution is
\begin{equation}
\label{eq:conditional-distribution}
\hat p_t^c=\frac{1}{N_t^c}\sum_{s\in\mathcal I_t^c}h(Y_s),
\qquad N_t^c>0.
\end{equation}
As a practical control target, we relax Equation~\ref{eq:fairness} to an \emph{acceptable fairness target set} $\mathcal F_c\subseteq\Delta^{K-1}$ for each concept $c$, assumed nonempty, closed, and convex, and to contain the prescribed target distribution $\pi$. Rather than requiring exact equality with $\pi$, the controller seeks to steer the empirical outcome distribution toward $\mathcal F_c$. The Euclidean projection onto $\mathcal F_c$ defines the correction residual:
\begin{equation}
\label{eq:residual}
q_t^c=\Pi_{\mathcal F_c}(\hat p_t^c),
\qquad
g_t^c=\hat p_t^c-q_t^c.
\end{equation}
The residual vanishes on $\mathcal F_c$, and its norm measures the fairness gap. Fair Memory stores $\mathcal M_t(c)=(N_t^c,\hat p_t^c,g_t^c)$, providing the state for subsequent requests of the same concept. The target set $\mathcal F_c$ is fixed per concept and supplied to the controller as an input alongside $\pi$.

\subsection{Fairness-Aware Prompt Optimization}
\label{sec:action-selection}
For an attribute-unspecified prompt $x_t$, fairness-aware prompt optimization translates the concept-specific fairness residual into an explicit attribute intervention in the input text.

\paragraph{Fairness-guided attribute selection.}
For a concept with existing Fair Memory, the controller selects the attribute that minimizes the fairness residual score:
\begin{equation}
\label{eq:approachability-action}
a_t^\star = \arg\min_{a\in\mathcal A}
\langle g_{t-1}^{c_t}, h(a)\rangle.
\end{equation}
This rule favors attributes that are underrepresented relative to the projected fairness target. When multiple attributes attain the same minimum score, ties are broken by sampling according to their relative probabilities in the target distribution $\pi$. If the empirical distribution is already within $\mathcal F_c$, then $g_{t-1}^{c_t}=0$, all attributes have equal scores, and the controller samples from $\pi$.
For a new concept ($N_{t-1}^{c_t}=0$), the controller samples $a_t\sim\pi$ before applying the same optimizing procedure.

For example, use male/female coordinates with $\pi=(0.5,0.5)$ and $\mathcal F_c=\{(u,1-u):u\in[0.4,0.6]\}$. Suppose $\hat p_{t-1}^{c_t}=(0.75,0.25)$. Since the male proportion exceeds the upper bound $0.6$, the closest acceptable distribution is $q_{t-1}^{c_t}=(0.6,0.4)$. Equation~\ref{eq:residual} then gives the residual scores $g_{t-1}^{c_t}=(0.15,-0.15)$. Equation~\ref{eq:approachability-action} therefore selects female, whose negative residual indicates underrepresentation relative to the projected target.

\paragraph{Semantics-preserving prompt optimization.}
The selected attribute $a_t$ is incorporated into the original prompt by an LLM optimizer:
\[
\tilde x_t=\mathcal R(x_t,a_t).
\]
Using a fixed instruction template, the LLM optimizer is instructed to make the selected attribute explicit for the target person while preserving the original concept, action, and scene context. For example, selecting female can turn ``Generate a video of a doctor rushing through a hospital corridor.'' into ``Generate a video of a \textit{female} doctor rushing through a hospital corridor.'' The optimized prompt is then passed to the video generator.

\paragraph{Outcome feedback and memory update.}
The generator produces $v_t=G(\tilde x_t)$, and an external classifier observes $Y_t=D(v_t)$. If the detection is invalid, the video is returned without updating Fair Memory. Otherwise, since the generator may not realize the selected attribute, memory is updated with the detected outcome $h(Y_t)$: increment $N_t^{c_t}$, recompute $\hat p_t^{c_t}$ using Equation~\ref{eq:conditional-distribution}, and update the residual. Other concepts' states remain unchanged. This feedback uses the requested video itself, requiring no separate calibration generations. Algorithm~\ref{alg:mem2fair} summarizes the procedure, updating Fair Memory in place.

\begin{algorithm}[t]
\caption{\method: one request $t$}
\label{alg:mem2fair}
\small
\begin{algorithmic}[1]
\Require prompt $x_t$; Fair Memory $\mathcal M$; target $\pi$; target sets $\{\mathcal F_c\}_c$
\Ensure video $v_t$ and updated Fair Memory $\mathcal M$
\If{$x_t$ specifies the protected attribute}
  \State \Return $(G(x_t),\mathcal M)$
\EndIf
\State $c_t\gets C(x_t)$
\If{$c_t$ has no valid observations in $\mathcal M$}
  \State sample $a_t\sim\pi$
\Else
  \State choose $a_t\in\arg\min_{a\in\mathcal A}\langle g^{c_t},h(a)\rangle$
\EndIf
\State $v_t\gets G(\mathcal R(x_t,a_t))$; $Y_t\gets D(v_t)$
\If{$Y_t\in\mathcal A$}
  \State update $N^{c_t}$ and $\hat p^{c_t}$ with $h(Y_t)$
  \State recompute $g^{c_t}$ using Equation~\ref{eq:residual}
  \State store $\mathcal M(c_t)\gets(N^{c_t},\hat p^{c_t},g^{c_t})$
\EndIf
\State \Return $(v_t,\mathcal M)$
\end{algorithmic}
\end{algorithm}

\begin{table*}[t]
\centering
\small
\setlength{\tabcolsep}{6pt}
\renewcommand{\arraystretch}{1.18}
\caption{Gender results on the 50 in-distribution occupations under the uniform target distribution $\pi=(\pi_{\text{male}},\pi_{\text{female}})=(1/2,1/2)$. Avg. is the arithmetic mean of the three quality metrics (Subj., Motion, and Imaging). \textbf{Bold} and \underline{underlined} means indicate the best and second-best results across all methods, respectively.}
\label{tab:main-results}
\begin{tabular}{lccccccc}
\toprule
\multirow{2}{*}{\textbf{Method}} & \multicolumn{2}{c}{\textbf{Fairness}} & \multicolumn{1}{c}{\textbf{Alignment}} & \multicolumn{4}{c}{\textbf{Quality}} \\
 & \emph{NP}$\uparrow$ & \emph{Bias}$\downarrow$ & \emph{CLIPsim}$\uparrow$ & \emph{Subj.}$\uparrow$ & \emph{Motion}$\uparrow$ & \emph{Imaging}$\uparrow$ & \emph{Avg.}$\uparrow$ \\
\midrule
\rowcolor[gray]{0.93}
Vanilla    & 0.336 & 0.384 & \textbf{0.1984}\pmstd{0.0279} & \underline{0.9815}\pmstd{0.0140} & \textbf{0.9958}\pmstd{0.0043} & \textbf{0.7430}\pmstd{0.0287} & \textbf{0.9068} \\
\rowcolor[gray]{0.97}
FairGen    & \underline{0.360} & 0.500 & 0.1706\pmstd{0.0361} & 0.9565\pmstd{0.0131} & 0.9731\pmstd{0.0034} & 0.6250\pmstd{0.0359} & 0.8515 \\
\rowcolor[gray]{0.97}
FairImagen & 0.287 & \underline{0.367} & 0.1742\pmstd{0.0392} & 0.9735\pmstd{0.0128} & 0.9752\pmstd{0.0040} & 0.7118\pmstd{0.0344} & 0.8868 \\
\rowcolor[gray]{0.97}
LightFair  & 0.342 & 0.369 & 0.1885\pmstd{0.0357} & 0.9802\pmstd{0.0119} & \underline{0.9907}\pmstd{0.0031} & 0.7025\pmstd{0.0371} & 0.8911 \\
\rowcolor[gray]{0.86}
\method    & \textbf{0.917} & \textbf{0.052} & \underline{0.1941}\pmstd{0.0301} & \textbf{0.9840}\pmstd{0.0121} & 0.9825\pmstd{0.0035} & \underline{0.7260}\pmstd{0.0322} & \underline{0.8975} \\
\bottomrule
\end{tabular}
\end{table*}

\subsection{Conditional Fairness Guarantee}
\label{sec:conditional-guarantee}

Fix a concept $c$ and index its valid detected outcomes by $n$. Let $Z_n^c=h(Y_n^c)$, and denote the corresponding empirical distribution, projection, and residual by $\hat p_n^c$, $q_n^c$, and $g_n^c$, respectively. Let $\mathcal H_{n-1}^c$ denote the history available before the $n$-th decision, including the current prompt.

The first valid observation initializes Fair Memory, so the directional condition is imposed from the second valid observation onward.

\begin{assumption}[Approximate directional condition]
\label{ass:approx-approachability}
For every concept $c$ and local round $n\geq2$,
\begin{equation}
\mathbb E\!\left[
\langle g_{n-1}^c,Z_n^c-q_{n-1}^c\rangle
\mid\mathcal H_{n-1}^c
\right]\leq\xi_c,
\end{equation}
where $\xi_c\geq0$ is the directional slack.
\end{assumption}

\begin{theorem}[Approximate conditional approachability]
\label{thm:approx-approachability}
Under Assumption~\ref{ass:approx-approachability}, after $n$ valid detected outcomes,
\begin{equation}
\mathbb E\!\left[
\operatorname{dist}^2(\hat p_n^c,\mathcal F_c)
\right]
\leq \xi_c+\frac{B^2}{n},
\end{equation}
where $B=\sqrt{2}$ is valid on the probability simplex.
\end{theorem}

The bound separates directional slack from finite-sample error, yielding an expected distance of at most $\sqrt{\xi_c}+B/\sqrt{n}$. Under perfect attribute compliance, the projection property and $\pi\in\mathcal F_c$ ensure that a corrective direction always exists, giving $\xi_c=0$ and an $O(n^{-1/2})$ convergence rate. Imperfect compliance may introduce nonzero slack, leaving an asymptotic upper bound of $\sqrt{\xi_c}$ on the expected distance.

The guarantee applies to each concept's detected-outcome distribution, including the cold-start observation. The proof and the relationship between directional slack and attribute selection are provided in Appendix~\ref{app:theory-proofs}.

\section{Experiments}
\label{sec:experiments}

\subsection{Experimental Setup}
\paragraph{Model and data.}

We use Wan2.2-5B as the base text-to-video generator~\citep{wan2025}. Our occupation set is drawn from the Stable Bias dataset~\citep{luccioni2023stablebias}. We evaluate three protected-attribute tasks: gender ($\{\text{female},\text{male}\}$), race ($\{\text{white},\allowbreak\text{black},\allowbreak\text{asian},\allowbreak\text{indian},\allowbreak\text{middle eastern},\allowbreak\text{latino hispanic}\}$), and age ($\{\text{young},\text{middle-aged},\text{old}\}$).

\paragraph{Compared methods}
Vanilla uses the unmodified Wan2.2 model as the reference. We apply FairGen~\citep{fairgen2025iccv}, FairImagen~\citep{fairimagen2025}, and LightFair~\citep{lightfair2025} to the same Wan2.2 backbone for comparison.

\paragraph{Metrics.}
Let $\mathcal{P}=\{x_n\}_{n=1}^{N}$ be an evaluation prompt set, let $S$ be the number of videos generated per prompt, let $S_n\leq S$ be the number with a valid attribute detection for prompt $n$, and let $Y_{n,s}\in\{a_1,\ldots,a_K\}$ be the detected attribute of valid video $s$ for prompt $n$. For a prescribed target distribution $\pi=(\pi_1,\ldots,\pi_K)\in\Delta^{K-1}$, define
\begin{equation}
\label{eq:experiment-proportions}
\begin{gathered}
p_i={\displaystyle\sum_{n=1}^{N}\sum_{s=1}^{S_n}\mathbb{I}\{Y_{n,s}=a_i\}} / {\displaystyle\sum_{n=1}^{N}S_n},\\[6pt]
p_i^{(n)}=\frac{1}{S_n}\sum_{s=1}^{S_n}\mathbb{I}\{Y_{n,s}=a_i\}.
\end{gathered}
\end{equation}
The \emph{Normalized Parity} (NP) pools all generated videos and measures the deviation from $\pi$, following the normalized-discrepancy formulation of \citet{teo2023measuringfairness} and its uniform-target instantiation in FairImagen~\citep{fairimagen2025}, which we generalize from a uniform reference to an arbitrary target $\pi$:
\begin{equation}
\label{eq:normalized-parity}
\mathrm{NP}_{\pi}
=1-\frac{\frac{1}{2}\sum_{i=1}^{K}|p_i-\pi_i|}
{1-\min_{i}\pi_i}\in[0,1].
\end{equation}
Here $\mathrm{NP}_{\pi}=1$ denotes an exact match to the target distribution, and larger values indicate greater global fairness. Detailed derivation in Appendix~\ref{app:np-derivation}. A prompt with $S_n=0$ contributes no attribute information, so we exclude it from per-prompt statistics; the \emph{Bias Score} therefore averages over the $N_{+}=|\{n:S_n>0\}|$ prompts with at least one valid detection:
\begin{equation}
\label{eq:bias-score}
\mathrm{Bias}_{\pi}
=\frac{1}{N_{+}}\sum_{n:\,S_n>0}
\frac{1}{K}\sum_{i=1}^{K}|p_i^{(n)}-\pi_i|
\in[0,1].
\end{equation}
Lower Bias Score indicates that individual prompts produce distributions closer to $\pi$. 
We also report \emph{CLIPsim} as prompt--video semantic alignment, computed by comparing the prompt representation with the visual representation of the video's middle frame:

\begin{equation}
\label{eq:clipsim}
\begin{aligned}
\mathrm{CLIPsim}
&=\frac{1}{NS}\sum_{n=1}^{N}\sum_{s=1}^{S}\\
&\quad\cdot\cos\!\left(\phi_{\mathrm{prompt}}(x_n),
\phi_{\mathrm{visual}}(v_{n,s}^{\mathrm{mid}})\right).
\end{aligned}
\end{equation}
Finally, we report the VBench~\citep{Huang_2024_VBench_CVPR} dimensions \texttt{subject\_consistency}, \texttt{motion\_smoothness}, and \texttt{imaging\_quality}, for which higher values are better.

\begin{figure*}[t]
\centering
\includegraphics[width=\textwidth]{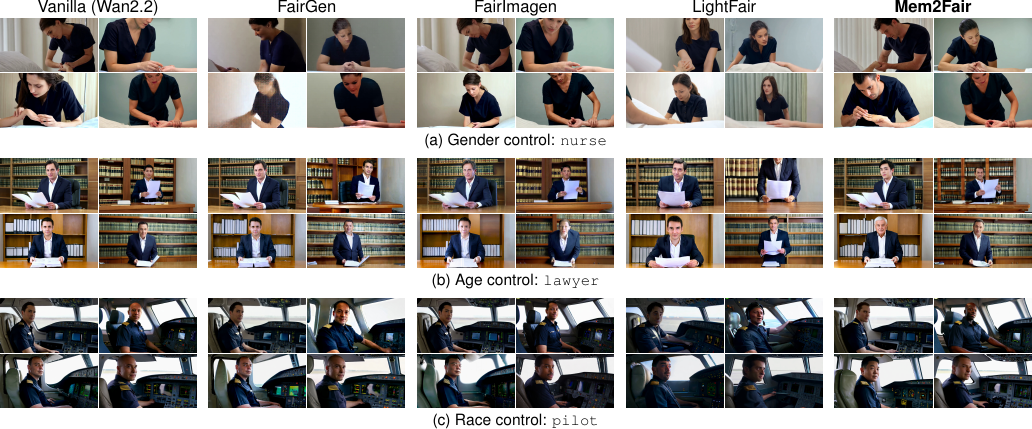}
\captionsetup{skip=3pt}
\caption{Early demographic representation under uniform target distributions. Each $2\times2$ group shows the first four consecutive generations for a concept and method, with one representative frame per video. \method{} starts each stream with empty Fair Memory.}
\label{fig:qualitative-results}
\end{figure*}

\begin{figure}[t]
\centering
\includegraphics[width=\columnwidth]{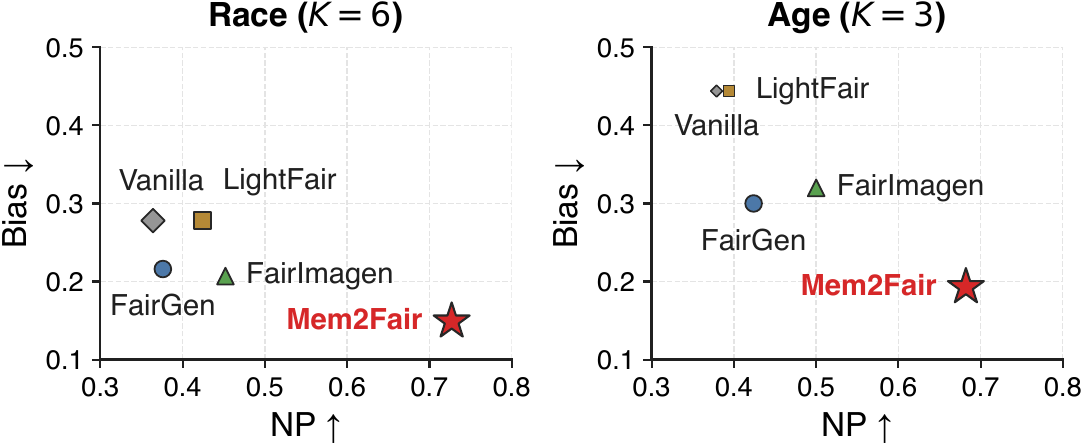}
\caption{Race ($K=6$, left) and age ($K=3$, right) results on the 50 seen occupation concepts under uniform target distributions. Higher NP and lower Bias are better. Red stars denote \method{}.}
\label{fig:race-age-results}
\end{figure}

\paragraph{Implementation Details.}
We use Qwen3-0.6B to map each prompt to an occupation concept by matching it against a predefined occupation vocabulary, and to rewrite prompts using a fixed instruction template. DeepFace~\citep{serengil2020deepface} provides the output-side attribute classifier, with RetinaFace~\citep{deng2020retinaface} as the face detection backend. Wan2.2 generates videos with 49 frames at a resolution of $1280 \times 704$. Unless otherwise specified, all settings use one test prompt per occupation and eight independent random seeds per prompt, with the same prompts and seed protocol across methods. Fair Memory is initialized from scratch for each experimental setting.
We compute CLIPsim using CLIP \texttt{ViT-L/14}.
Across all attributes, concepts, and target distributions, we use a normalized TV tolerance $\alpha=0.2$:
\[
\mathcal F_c=\left\{p\in\Delta^{K-1}:\frac{\tfrac12\|p-\pi\|_1}{1-\min_i\pi_i}\le\alpha\right\},
\]
with Euclidean projection onto $\mathcal F_c$.

\subsection{Main Results}
\label{sec:main-results}

\paragraph{In-Distribution Gender Control.}
We evaluate gender control on 50 seen occupation concepts under the uniform target $\pi=(1/2,1/2)$.

Table~\ref{tab:main-results} compares \method{} with the unmodified generator and baselines. \method{} raises NP from $0.336$ to $0.917$ and lowers Bias Score from $0.384$ to $0.052$, a larger margin than any baseline achieves. \method{} also exceeds all three fairness baselines in CLIPsim (0.1941) and average quality (0.8975).

\begin{figure*}[t]
\centering
\includegraphics[width=\textwidth]{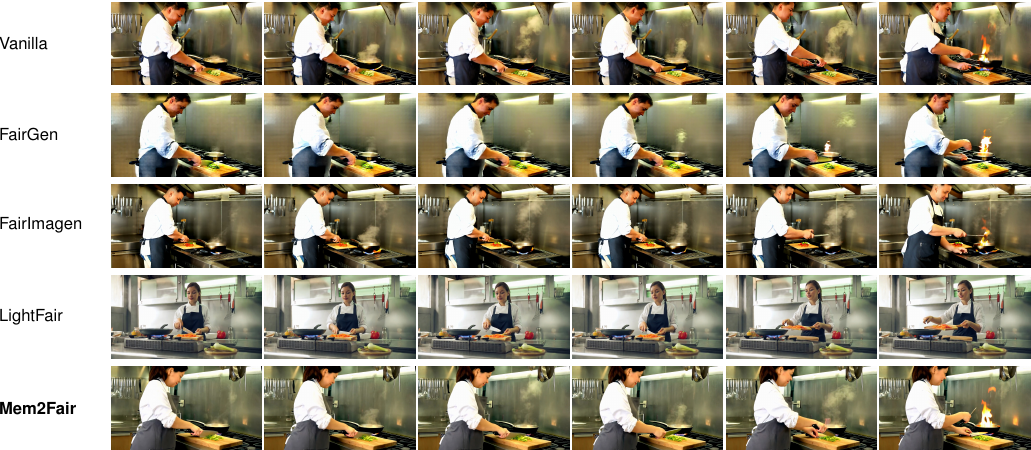}
\captionsetup{skip=3pt}
\caption{Video quality comparison for \texttt{cook}. Each row shows one generated video; columns show frames in chronological order from left to right, with timestamps matched across methods.}
\label{fig:quality-cook}
\end{figure*}

FairGen incurs the largest quality loss, particularly in imaging quality. FairImagen also reduces CLIPsim and average video quality relative to Vanilla. Changing denoising or text embeddings may also affect visual details and motion consistency in Wan2.2. We also provide a qualitative video quality comparison in Figure~\ref{fig:quality-cook}.

LightFair preserves relatively high video quality but provides limited fairness correction in this setting. This may reflect a limitation of transferring its two-stage guidance scheme to Wan2.2, where restricting debiased conditioning to selected generation stages may provide insufficient control over demographic outcomes.

\paragraph{Multi-Attribute Control on Race and Age.}
We evaluate race ($K=6$) and age ($K=3$) on the same 50 seen occupation concepts with a uniform target for each attribute.

Figure~\ref{fig:race-age-results} reports the results. \method{} achieves the highest NP and lowest Bias Score for both attributes relative to the strongest baseline.
Figure~\ref{fig:qualitative-results} illustrates early demographic representation across consecutive generations for the example concepts, complementing the aggregate distributional results.

\begin{table*}[t]
\centering
\small
\setlength{\tabcolsep}{6pt}
\renewcommand{\arraystretch}{1.18}
\caption{Generalization and robustness results for gender under the uniform target distribution $\pi=(\pi_{\text{male}},\pi_{\text{female}})=(1/2,1/2)$. \emph{Unseen Concept}: 25 held-out occupations with simple prompts. \emph{Complex Prompt}: the 50 occupations of the Main setting, queried with richer action and contextual descriptions. Avg. is the arithmetic mean of the three quality metrics (Subj., Motion, and Imaging). \textbf{Bold} and \underline{underlined} means indicate the best and second-best results within each setting, respectively.}
\label{tab:generalization}
\begin{tabular}{llccccccc}
\toprule
\multirow{2}{*}{\textbf{Setting}} & \multirow{2}{*}{\textbf{Method}} & \multicolumn{2}{c}{\textbf{Fairness}} & \multicolumn{1}{c}{\textbf{Alignment}} & \multicolumn{4}{c}{\textbf{Quality}} \\
 &  & \emph{NP}$\uparrow$ & \emph{Bias}$\downarrow$ & \emph{CLIPsim}$\uparrow$ & \emph{Subj.}$\uparrow$ & \emph{Motion}$\uparrow$ & \emph{Imaging}$\uparrow$ & \emph{Avg.}$\uparrow$ \\
\midrule
\rowcolor[gray]{0.93}
\cellcolor{white} & Vanilla    & \underline{0.290} & \underline{0.375} & \textbf{0.1919}\pmstd{0.0355} & \textbf{0.9814}\pmstd{0.0124} & \textbf{0.9926}\pmstd{0.0042} & \textbf{0.7380}\pmstd{0.0411} & \textbf{0.9040} \\
\rowcolor[gray]{0.97}
\cellcolor{white} & FairGen    & \underline{0.290} & 0.500 & \underline{0.1892}\pmstd{0.0347} & 0.9481\pmstd{0.0116} & 0.9801\pmstd{0.0021} & 0.6431\pmstd{0.0359} & 0.8571 \\
\rowcolor[gray]{0.97}
\cellcolor{white} & FairImagen & 0.250 & 0.385 & 0.1839\pmstd{0.0378} & 0.9708\pmstd{0.0108} & 0.9721\pmstd{0.0047} & 0.7069\pmstd{0.0446} & 0.8833 \\
\rowcolor[gray]{0.97}
\cellcolor{white} & LightFair  & 0.200 & 0.500 & 0.1876\pmstd{0.0362} & \underline{0.9811}\pmstd{0.0089} & 0.9853\pmstd{0.0033} & \underline{0.7174}\pmstd{0.0358} & \underline{0.8946} \\
\rowcolor[gray]{0.86}
\cellcolor{white}\multirow{-5}{*}{Unseen Concept} & \method    & \textbf{0.693} & \textbf{0.173} & 0.1889\pmstd{0.0378} & 0.9746\pmstd{0.0122} & \underline{0.9913}\pmstd{0.0033} & 0.7100\pmstd{0.0407} & 0.8920 \\
\midrule
\rowcolor[gray]{0.93}
\cellcolor{white} & Vanilla    & \underline{0.319} & 0.388 & \textbf{0.2515}\pmstd{0.0333} & \textbf{0.9738}\pmstd{0.0202} & \textbf{0.9964}\pmstd{0.0046} & \textbf{0.7368}\pmstd{0.0419} & \textbf{0.9023} \\
\rowcolor[gray]{0.97}
\cellcolor{white} & FairGen    & 0.267 & 0.500 & 0.2409\pmstd{0.0355} & 0.9502\pmstd{0.0211} & \underline{0.9896}\pmstd{0.0038} & 0.6019\pmstd{0.0421} & 0.8472 \\
\rowcolor[gray]{0.97}
\cellcolor{white} & FairImagen & 0.306 & \underline{0.362} & \underline{0.2451}\pmstd{0.0331} & 0.9631\pmstd{0.0208} & 0.9858\pmstd{0.0043} & 0.6756\pmstd{0.0458} & 0.8748 \\
\rowcolor[gray]{0.97}
\cellcolor{white} & LightFair  & 0.227 & 0.500 & 0.2316\pmstd{0.0338} & \underline{0.9705}\pmstd{0.0199} & 0.9661\pmstd{0.0037} & 0.7061\pmstd{0.0407} & 0.8809 \\
\rowcolor[gray]{0.86}
\cellcolor{white}\multirow{-5}{*}{Complex Prompt} & \method    & \textbf{0.856} & \textbf{0.075} & 0.2427\pmstd{0.0405} & 0.9611\pmstd{0.0181} & 0.9877\pmstd{0.0035} & \underline{0.7271}\pmstd{0.0295} & \underline{0.8920} \\
\bottomrule
\end{tabular}
\end{table*}

\paragraph{Generalization to Unseen Occupations.}
The \emph{Unseen Concept} setting evaluates gender control on 25 held-out occupations under the uniform target.

In Table~\ref{tab:generalization}, \method{} achieves the best NP $=0.693$ and Bias Score $=0.173$. The baselines show limited fairness transfer to unseen occupations, yielding even higher Bias Scores. This is likely because FairGen’s sampled attribute adapters and LightFair’s fixed two-stage guidance do not adjust to each occupation’s realized output history. Under concept shift, this may leave occupation-specific control errors uncorrected, allowing within-occupation imbalance to persist or worsen. Mem2Fair instead uses concept-specific outcome feedback to steer subsequent generations toward the target.

\paragraph{Robustness to Complex Prompts.}
The \emph{Complex Prompt} setting evaluates the same 50 concepts as the in-distribution setting under the uniform target, with prompts enriched by occupation-specific actions and contextual descriptions. Methods requiring training or calibration use corresponding complex prompts during preparation.

As reported in Table~\ref{tab:generalization}, \method{} attains the strongest fairness results in this setting, with the highest average quality among baselines. FairImagen retains relatively strong prompt alignment but offers limited fairness correction. This may be because suppressing demographic directions in the embedding space does not directly enforce balanced demographic outcomes in the video generator.

\begin{table}[t]
\centering
\small
\setlength{\tabcolsep}{3pt}
\caption{Non-Uniform Target Control on the 50 seen occupation concepts under $\pi=(\pi_{\text{male}},\pi_{\text{female}})=(1/4,3/4)$. Average scores are reported in the Quality group. FairImagen does not support non-uniform targets.}
\label{tab:controllable-results}
\begin{tabular}{lcc|c|c}
\toprule
\multirow{2}{*}{Method} & \multicolumn{2}{c|}{\emph{Fairness}} & \multicolumn{1}{c|}{\emph{Alignment}} & \emph{Quality} \\
\cmidrule(lr){2-3}\cmidrule(lr){4-4}\cmidrule(lr){5-5}
 & NP$\uparrow$ & Bias$\downarrow$ & CLIPsim$\uparrow$ & Avg.$\uparrow$ \\
\midrule
Vanilla    & 0.193 & 0.605 & 0.1991 & 0.8982 \\
FairGen    & 0.193 & 0.678 & 0.1902 & 0.8310 \\
FairImagen & \na & \na & \na & \na \\
LightFair  & 0.204 & 0.661 & 0.1873 & 0.8919 \\
\method    & \textbf{0.871} & \textbf{0.097} & 0.1914 & 0.8846 \\
\bottomrule
\end{tabular}
\end{table}

\begin{figure}[t]
\centering
\includegraphics[width=\columnwidth]{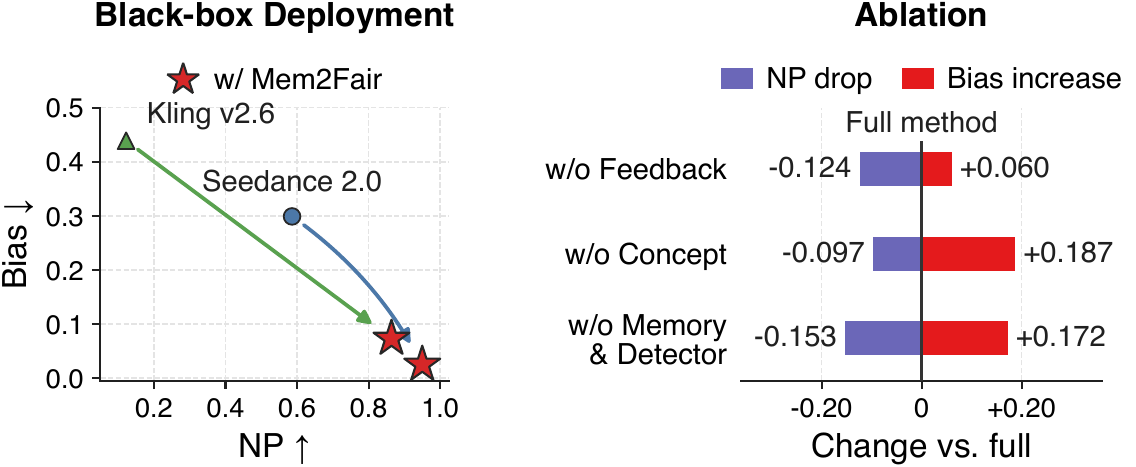}
\caption{\textbf{Left:} Black-box fairness on Seedance 2.0 and Kling v2.6 under the uniform gender target. Arrows connect Original outputs to outputs with \method{} (red stars). \textbf{Right:} Ablation changes relative to the full method (NP $=0.917$, Bias $=0.052$). The zero line denotes the full method.}
\label{fig:blackbox-ablation}
\end{figure}

\paragraph{Non-Uniform Target Control.}
We evaluate gender control on the same 50 seen occupation concepts with the non-uniform target $\pi=(\pi_{\text{male}},\pi_{\text{female}})=(1/4,3/4)$. FairImagen does not support non-uniform targets and is omitted from this comparison.

Table~\ref{tab:controllable-results} shows that \method{} follows the prescribed target more closely than the supported baselines, reaching NP $=0.871$ and Bias Score $=0.097$ while preserving prompt-video alignment and imaging quality. This supports the controller's ability to track prescribed demographic proportions beyond uniform parity, consistent with computing corrections relative to the target-dependent fairness set.

\paragraph{Black-Box Generator Deployment.}
We evaluate \method{} on Seedance 2.0\footnote{Seedance 2.0 model homepage: \url{https://seed.bytedance.com/en/seedance2_0}.} and Kling v2.6\footnote{Kling AI model homepage: \url{https://klingai.com/global/}.} using the same 50 seen occupation concepts under the uniform target. The APIs are accessed only through inputs and outputs, without inspecting model parameters or internal representations.

Figure~\ref{fig:blackbox-ablation} (left) shows that \method{} improves fairness on both APIs: on Seedance 2.0, NP rises from $0.586$ to $0.950$ and Bias Score falls from $0.299$ to $0.025$; on Kling v2.6, NP rises from $0.122$ to $0.864$ and Bias Score falls from $0.439$ to $0.073$. The gains across APIs with different initial imbalances suggest that outcome feedback can support generator-specific correction using only input/output access.

\subsection{Ablation Study}
\label{sec:ablation}
We evaluate the full method and three ablated variants, each defined as follows. \emph{Full method} uses a separate Fair Memory for each concept and updates it with the attribute detected from each generated video. \emph{w/o Outcome Feedback} updates the memory with the requested attribute rather than the attribute detected from the generated video. \emph{w/o Concept Conditioning} uses one shared memory across all concepts instead of maintaining concept-specific memories. \emph{w/o Memory \& Detector} directly samples an attribute from global target distribution and rewrites the prompt.

Figure~\ref{fig:blackbox-ablation} (right) reports metric changes relative to the full method, which achieves NP $=0.917$ and Bias Score $=0.052$.
All ablated variants are worse than the full method. Sharing one memory across concepts still gives a moderate overall NP ($0.820$), but it increases the within-concept imbalance, producing the highest Bias Score among the ablated variants ($0.239$). Using the requested attribute as feedback gives NP $=0.793$ and Bias $=0.112$; when the generator realizes an attribute imperfectly, this variant records the request instead of the realized outcome and therefore cannot correct the resulting error. Removing both Memory and Detector gives the lowest ablation NP ($0.764$): direct sampling and rewriting ignores concept-specific history and realized-outcome feedback, leaving the generator's demographic bias uncorrected. Together, these results show that concept-specific state and realized-outcome feedback each contribute to fairness control.

\section{Conclusion}
\label{sec:conclusion}
We formulate fair text-to-video generation as a conditional online approachability problem and introduce \method, a plug-and-play framework that uses concept-conditioned Fair Memory for sequential fairness control. By operating solely through input prompts and generated videos, \method leaves the generator's internals and temporal process untouched. Our theoretical analysis establishes a finite-sample bound on the expected concept-level fairness discrepancy, accounting for finite observations and imperfect directional correction. Experiments demonstrate consistent fairness improvements across demographic attributes, prompt settings, and target distributions, with comparable prompt--video alignment and video quality, including on commercial black-box T2V models.

\section*{LLM Usage Statement}
LLMs were used as assistive tools in the experiments and manuscript preparation. In the experimental pipeline, Qwen3-0.6B assisted with mapping prompts to occupation concepts and rewriting prompts to incorporate the selected protected-attribute value using a fixed instruction template, as described in Section~\ref{sec:experiments}. OpenAI ChatGPT assisted with manuscript editing, language refinement and formatting. The authors retain responsibility for the research design, experimental execution, analysis, scientific claims, references, and final manuscript.

{
    \small
    \bibliographystyle{ieeenat_fullname}
    \bibliography{main}
}

\clearpage
\maketitlesupplementary
\appendix

\section{Additional Method Details and Analysis}
\label{sec:analysis}
\label{app:theory-proofs}

This appendix supplements Section~\ref{sec:method} with formal history definitions, the proof of the conditional fairness guarantee, and the analysis of attribute selection.

\subsection{History and Cold Start}
\label{app:method-details}
Throughout this appendix, all counts and sequences are restricted to attribute-unspecified requests with valid detected outcomes (as defined in Section~\ref{sec:fair-memory}); requests that are user-specified or yield an invalid detection are excluded from every index, count, and expectation below.
For a global request index $s$, let $c_s=C(x_s)$. Using the eligible index set $\mathcal I_t^c$ from Section~\ref{sec:fair-memory}, $N_t^c=|\mathcal I_t^c|$ counts only attribute-unspecified requests with valid detected outcomes. User-specified attribute values are passed through unchanged and do not update Fair Memory. For requests with an existing state,
\begin{equation}
\label{eq:memory-update}
\hat p_t^{c_t}
=\frac{N_{t-1}^{c_t}\hat p_{t-1}^{c_t}+h(Y_t)}
{N_{t-1}^{c_t}+1}.
\end{equation}
The projection used in Equation~\ref{eq:residual} is
\[
\Pi_{\mathcal F_c}(p)
=\arg\min_{q\in\mathcal F_c}\|p-q\|_2.
\]
It is unique because $\mathcal F_c$ is closed, convex, and nonempty.

For cold-start requests, the controller samples $a_t\sim\pi$. Because all $K$ attribute values are admissible, the distribution of selected attributes coincides with the target; the detected outcome is then a draw from the realized-outcome mixture $\sum_k\pi_k\mu_t(a_k)$, which equals $\pi$ only when the generator realizes every attribute exactly. The main theorem therefore does not rely on this coincidence and instead uses the directional condition on detected outcomes.

\subsection{Proof of Approximate Conditional Approachability}

\begin{proof}[Proof of Theorem~\ref{thm:approx-approachability}]
Fix a concept $c$. Let $B$ satisfy $\lVert z-q\rVert_2\leq B$ for every one-hot $z$ and $q\in\mathcal F_c$; $B=\sqrt{2}$ is valid on the simplex. Abbreviate
\begin{equation}
d_n=\operatorname{dist}(\hat p_n^c,\mathcal F_c).
\end{equation}
For each $n\geq2$, also write
\begin{equation}
q_{n-1}=\Pi_{\mathcal F_c}(\hat p_{n-1}^c),
\qquad
g_{n-1}=\hat p_{n-1}^c-q_{n-1}.
\end{equation}
The empirical update over the concept-specific subsequence is, for $n\geq2$,
\begin{equation}
\hat p_n^c
=\frac{n-1}{n}\hat p_{n-1}^c+\frac{1}{n}Z_n^c.
\end{equation}
Because $q_{n-1}\in\mathcal F_c$,
\begin{align}
d_n^2
&\leq\|\hat p_n^c-q_{n-1}\|_2^2\\
&=\frac{(n-1)^2}{n^2}d_{n-1}^2 \notag\\
&\quad+\frac{2(n-1)}{n^2}
\langle g_{n-1},Z_n^c-q_{n-1}\rangle \notag\\
&\quad+\frac{1}{n^2}\|Z_n^c-q_{n-1}\|_2^2.
\end{align}
Taking conditional expectation and applying Assumption~\ref{ass:approx-approachability} and the definition of $B$ gives
\begin{equation}
n^2\mathbb E[d_n^2]
\leq
(n-1)^2\mathbb E[d_{n-1}^2]
+2(n-1)\xi_c+B^2.
\end{equation}
For $n=1$, $d_1^2\leq B^2$. Summing the recursion from $2$ to $n$ therefore yields
\begin{equation}
n^2\mathbb E[d_n^2]
\leq n(n-1)\xi_c+nB^2.
\end{equation}
Therefore,
\begin{equation}
\mathbb E[d_n^2]
\leq\left(1-\frac{1}{n}\right)\xi_c+\frac{B^2}{n}
\leq\xi_c+\frac{B^2}{n}.
\end{equation}
The first outcome comes from the target-guided cold start, and the base case requires no directional condition. Jensen's inequality and $\sqrt{u+v}\leq\sqrt u+\sqrt v$ give the distance bound in the main text.
\end{proof}

\subsection{How Attribute Selection Enters the Guarantee}
\label{sec:controller-slack}

\paragraph{Nominal and true effects.}
The nominal effect used by the selector is the one-hot encoding of a candidate attribute:
\begin{equation}
\label{eq:nominal-action-effect}
\hat\mu_t(a)\mathrel{:=}h(a).
\end{equation}
Let $\mathcal H_{t-1}$ contain the interaction history before request $t$. The true effect denotes the conditional expected detected outcome when the prompt is rewritten with $a$:
\begin{equation}
\label{eq:true-action-effect}
\mu_t(a)
=\mathbb E\!\left[h(D(G(\mathcal R(x_t,a))))
\mid\mathcal H_{t-1},x_t,D\text{ valid}\right].
\end{equation}
This is an analytical quantity, not a separate predictor queried by the controller.

Theorem~\ref{thm:approx-approachability} uses a local index $n$, whereas the controller in Section~\ref{sec:method} uses the global request index $t$. At request $t$ for concept $c_t$ with an existing memory state, the projection and residual $q_{t-1}^{c_t}=\Pi_{\mathcal F_{c_t}}(\hat p_{t-1}^{c_t})$ and $g_{t-1}^{c_t}=\hat p_{t-1}^{c_t}-q_{t-1}^{c_t}$ are the local quantities $q_{n-1}^{c_t}$ and $g_{n-1}^{c_t}$ for that concept. The controller has only one decision: choose $a\in\mathcal A$ and rewrite the prompt with that attribute. In the notation of Section~\ref{sec:action-selection}, its nominal effect is $\hat\mu_t(a)=h(a)$, while its true effect after rewriting and generation is $\mu_t(a)$. The theorem does not assume that the discrete selector in Equation~\ref{eq:approachability-action} is solved exactly. To relate its slack to this selector, suppose that for each request after cold start and candidate attribute $a\in\mathcal A$,
\[
\lVert\hat\mu_t(a)-\mu_t(a)\rVert_2\leq\delta,
\qquad
\lVert g_{t-1}^{c_t}\rVert_2\leq G_g.
\]
Here $\delta$ bounds prompt-compliance error and $G_g$ bounds the residual norm ($G_g=\sqrt{2}$ is valid because both the empirical distribution and its projection lie in $\Delta^{K-1}$). Assume also that the attribute selector is $\eta$-optimal, meaning that the selected attribute satisfies
\[
\left\langle g_{t-1}^{c_t},h(a_t)\right\rangle
\leq\min_{a\in\mathcal A}\left\langle g_{t-1}^{c_t},h(a)\right\rangle+\eta .
\]
No reachability condition on the action set is needed: because $\pi\in\mathcal F_c$ and $q_{t-1}^{c_t}$ is the Euclidean projection of $\hat p_{t-1}^{c_t}$ onto a closed convex $\mathcal F_c$, the projection inequality gives $\langle g_{t-1}^{c_t},\pi-q_{t-1}^{c_t}\rangle\leq0$, and writing $\pi=\sum_{k}\pi_kh(a_k)$ yields
\[
\min_{a\in\mathcal A}\left\langle g_{t-1}^{c_t},h(a)-q_{t-1}^{c_t}\right\rangle
\leq\left\langle g_{t-1}^{c_t},\pi-q_{t-1}^{c_t}\right\rangle\leq0 .
\]
Hence at least one attribute always offers a nonpositive correction score, and the directional condition holds with
\[
\xi_c\leq G_g\delta+\eta.
\]
The two terms account for prompt-compliance error and approximate attribute selection, respectively. For the tie-breaking rule of Section~\ref{sec:action-selection}, every tied attribute attains the minimum score exactly, so $\eta=0$ and $\xi_c\leq G_g\delta$.

\paragraph{Derivation of the directional slack.}
The compliance bound holds conditionally on the pre-decision history. Inserting the nominal effect and applying Cauchy--Schwarz to the prompt-compliance error gives, for any $a\in\mathcal A$,
\begin{align}
&\left\langle g_{t-1}^{c_t},\mu_t(a)-q_{t-1}^{c_t}\right\rangle \notag\\
&\quad=\left\langle g_{t-1}^{c_t},h(a)-q_{t-1}^{c_t}\right\rangle \notag\\
&\qquad+\left\langle g_{t-1}^{c_t},\mu_t(a)-h(a)\right\rangle\\
&\quad\leq\left\langle g_{t-1}^{c_t},h(a)-q_{t-1}^{c_t}\right\rangle
+G_g\delta .
\end{align}
Choosing $a=a_t$ and using $\eta$-optimality together with the projection inequality above,
\begin{align}
&\left\langle g_{t-1}^{c_t},\mu_t(a_t)-q_{t-1}^{c_t}\right\rangle \notag\\
&\quad\leq\min_{a\in\mathcal A}\left\langle g_{t-1}^{c_t},h(a)-q_{t-1}^{c_t}\right\rangle \notag\\
&\qquad+G_g\delta+\eta\\
&\quad\leq\left\langle g_{t-1}^{c_t},\pi-q_{t-1}^{c_t}\right\rangle
+G_g\delta+\eta \notag\\
&\quad\leq G_g\delta+\eta .
\end{align}
Thus Assumption~\ref{ass:approx-approachability} holds with
\begin{equation}
\xi_c=G_g\delta+\eta .
\end{equation}
The compliance bound is an assumption about how reliably the generator realizes the selected attribute; it can be assessed from detected outcomes, without generating extra candidate videos.

\section{Derivation of the Normalized Parity Score}
\label{app:np-derivation}

This section derives the denominator of Equation~\ref{eq:normalized-parity} and relates $\mathrm{NP}_{\pi}$ to the fairness-discrepancy normalization of \citet{teo2023measuringfairness}.

\paragraph{Normalization principle.} \citet{teo2023measuringfairness} measure fairness by a discrepancy $f=D(\bar p,\mathbb E[C(x)])$ between a reference attribute distribution $\bar p$ and the distribution realized by the generator, and observe that the \emph{raw} discrepancy is not comparable across attributes with different numbers of groups $K$, because its maximum attainable value $f_{\max}$ grows with $K$. Their fix is to normalize by the discrepancy evaluated at the \emph{absolutely-biased} extreme point, i.e., $f/f_{\max}$, which places the score in $[0,1]$ regardless of $K$. FairImagen~\citep{fairimagen2025} instantiates this principle with the total-variation distance $D(p,\bar p)=\frac{1}{2}\sum_i|p_i-\bar p_i|$ and a uniform reference $\bar p_i=1/K$, for which the absolutely-biased extreme point is any one-hot vertex and $f_{\max}=1-1/K$. Our $\mathrm{NP}_{\pi}$ applies the same normalization principle with an arbitrary target $\pi$ in place of the uniform reference.

\paragraph{Maximum attainable deviation from a general target.} For a target $\pi$ on the $(K-1)$-simplex $\Delta^{K-1}$, define the total-variation deviation $\mathrm{TV}(p,\pi)=\frac{1}{2}\sum_{i=1}^K|p_i-\pi_i|$. We show
\begin{equation}
\label{eq:tv-max}
\max_{p\in\Delta^{K-1}}\mathrm{TV}(p,\pi)=1-\min_{i}\pi_i,
\end{equation}
attained at the one-hot vertex $p=e_{j}$ with $j=\arg\min_i \pi_i$. Since $\mathrm{TV}(\cdot,\pi)$ is a convex function of $p$ (a sum of absolute values of affine functions of $p$) and $\Delta^{K-1}$ is a polytope, its maximum over $\Delta^{K-1}$ is attained at a vertex, i.e., at some one-hot $e_j$. At any vertex $e_j$,
\begin{equation}
\begin{aligned}
\mathrm{TV}(e_j,\pi)
&=\frac{1}{2}\Big((1-\pi_j)+\sum_{i\neq j}\pi_i\Big)\\
&=\frac{1}{2}\big((1-\pi_j)+(1-\pi_j)\big)\\
&=1-\pi_j,
\end{aligned}
\end{equation}
using $\sum_{i\neq j}\pi_i=1-\pi_j$. This is maximized by choosing $j=\arg\min_i\pi_i$, giving Equation~\ref{eq:tv-max}. Substituting $f=\mathrm{TV}(p,\pi)$ and $f_{\max}=1-\min_i\pi_i$ into the normalization $1-f/f_{\max}$ of \citet{teo2023measuringfairness} recovers Equation~\ref{eq:normalized-parity} exactly, with the sign convention flipped so that $\mathrm{NP}_{\pi}=1$ (rather than $0$) denotes the fairest outcome.

\paragraph{Recovery of FairImagen and semantics of the uniform case.} When $\pi_i=1/K$ for all $i$, Equation~\ref{eq:tv-max} gives $f_{\max}=1-1/K$ and every one-hot vertex attains it, so the worst case is exactly "all generations collapse onto a single group,'' matching the original FairImagen parity score. For a non-uniform $\pi$, the vertices are no longer equidistant from the target: collapsing onto the already-favored group ($\arg\max_i\pi_i$) yields a strictly smaller deviation than collapsing onto the least-favored group. Because $f_{\max}$ is defined by the worst case, $\mathrm{NP}_{\pi}=0$ always denotes collapse onto $\arg\min_i\pi_i$, and intermediate values of $\mathrm{NP}_{\pi}$ should be read as a fraction of this worst-case deviation from $\pi$ rather than as a fraction of "distance from a single-group outcome'' in general.

\section{Experimental Configuration}
\label{app:experimental-configuration}

This section supplements Section~\ref{sec:experiments} with prompt preparation details and detection coverage.

\paragraph{Prompt preparation.}
Methods requiring training or calibration use the $50$ seen occupations, disjoint from the $25$ held-out occupations. Complex prompts are constructed from O*NET 31.0 task descriptions before evaluation and fixed across methods, without runtime regeneration or resampling.

\paragraph{Valid detection coverage.}
\label{app:coverage}
For method $m$ and protected attribute $\mathcal A$, we define detection coverage using the counts in Equation~\ref{eq:experiment-proportions}:
\begin{equation}
\label{eq:coverage}
\operatorname{Coverage}(m,\mathcal A)
=\frac{\sum_{n=1}^{N}S_{n}}{N\cdot S}.
\end{equation}
Failed detections remain in the denominator. Table~\ref{tab:coverage} reports coverage in the Main setting. If no video yields a valid detection ($N_{+}=0$), both fairness metrics are undefined and reported as dashes.

\begin{table}[!htbp]
\centering
\small
\setlength{\tabcolsep}{4pt}
\caption{Detection coverage in the Main setting: validly detected videos / generated videos, for each method and attribute task.}
\label{tab:coverage}
\begin{tabular}{lccc}
\toprule
Method & Gender ($K=2$) & Race ($K=6$) & Age ($K=3$) \\
\midrule
Vanilla    & 399/400 & 396/400 & 396/400 \\
FairGen    & 400/400 & 396/400 & 396/400 \\
FairImagen & 397/400 & 400/400 & 390/400 \\
LightFair  & 400/400 & 396/400 & 396/400 \\
\method    & 397/400 & 396/400 & 396/400 \\
\bottomrule
\end{tabular}
\end{table}

\end{document}